\documentclass[11pt,a4paper]{article}
\usepackage{fullpage}
\usepackage{booktabs} 
\usepackage{subfigure}
\usepackage{amsmath}
\usepackage{amssymb}
\usepackage{mathtools}
\usepackage{amsthm}
\usepackage{url}

\theoremstyle{plain}
\newtheorem{theorem}{Theorem}[section]

\newtheorem{corollary}[theorem]{Corollary}
\theoremstyle{definition}

\theoremstyle{remark}

\newcommand{\oms}{\{\!\!\{}
\newcommand{\cms}{\}\!\!\}}

\newcommand{\bigoms}{\big\{\!\!\big\{}
\newcommand{\bigcms}{\big\}\!\!\big\}}

\newcommand{\eg}{e.\,g., }
\newcommand{\ie}{i.\,e., }


\title{\textbf{Graph Neural Machine: A New Model for Learning with Tabular Data}}

\author{Giannis Nikolentzos$^1$ \\ \texttt{nikolentzos@uop.gr} \and Siyun Wang$^2$ \\ \texttt{siyun.wang@polytechnique.edu} \and Johannes Lutzeyer$^2$ \\ \texttt{johannes.lutzeyer@polytechnique.edu} \and Michalis Vazirgiannis$^2$ \\ \texttt{mvazirg@lix.polytechnique.fr}}
\date{%
    $^1$University of Peloponnese, Greece\\%
    $^2$LIX, \'Ecole Polytechnique, IP Paris, France
}

\begin{document}
\maketitle 

\begin{abstract}
In recent years, there has been a growing interest in mapping data from different domains to graph structures.
Among others, neural network models such as the multi-layer perceptron (MLP) can be modeled as graphs.
In fact, MLPs can be represented as directed acyclic graphs.
Graph neural networks (GNNs) have recently become the standard tool for performing machine learning tasks on graphs.
In this work, we show that an MLP is equivalent to an asynchronous message passing GNN model which operates on the MLP's graph representation.
We then propose a new machine learning model for tabular data, the so-called Graph Neural Machine (GNM), which replaces the MLP's directed acyclic graph with a nearly complete graph and which employs a synchronous message passing scheme.
We show that a single GNM model can simulate multiple MLP models.
We evaluate the proposed model in several classification and regression datasets.
In most cases, the GNM model outperforms the MLP architecture.
\end{abstract}

\section{Introduction}
Graphs arise naturally in a wide range of domains, such as in social networks, in bio-informatics, and in chemo-informatics.
Learning useful representations from graph data is essential for many real-world applications.
For instance, in social network analysis, one might be interested in predicting the interests of users represented by the nodes of a network~\cite{yang2011like}.
In biology, an issue of high interest is the prediction of the functions of proteins modeled as graphs~\cite{gligorijevic2021structure}.
This has motivated the development of machine learning algorithms which can process graphs and extract useful representations.

In recent years, graph neural networks (GNNs) have achieved significant success in the field of graph representation learning~\cite{zhou2020graph}.
While there exist different families of GNNs (\eg message passing models~\cite{gilmer2017neural}, graph transformers~\cite{ying2021transformers}), message passing models are the most-studied GNN models and have seen several real-world applications~\cite{gori2005new,scarselli2009graph,li2015gated,hamilton2017inductive,kipf2017semi,velickovic2018graph,zhang2018end,xu2018powerful,morris2018weisfeiler}.
All these models share the same basic idea and can be reformulated into a single common framework.
Specifically, those models employ a message passing procedure, where each node updates its feature vector by aggregating the feature vectors of its neighbors~\cite{gilmer2017neural}.
After $k$ iterations of the message passing procedure, each node obtains a feature vector which captures the structural and feature information within its $k$-hop neighborhood.
These representations can be used as features for node-level tasks.
For graph-level tasks, GNNs typically compute a feature vector for the entire graph using some permutation invariant readout function which aggregates the representations of all the nodes.

Graphs are very general data structures, and therefore, several different types of data can be modeled as graphs. 
These types range from social networks and molecules to neural network models.
Indeed, we can represent a multi-layer perceptron (MLP) as a directed acyclic graph where nodes correspond to neurons in the neural network and edges correspond to connections between neurons.
In fact, this representation allows us to derive a connection between MLPs and GNNs.
A forward pass of an MLP is equivalent to a forward pass of an asynchronous message passing GNN on the aforementioned graph.
This graph representation of neural networks opens new directions for the design of neural architectures by leveraging the well-developed literature of GNNs.
More importantly, we can design new machine learning models by dropping the constraint of acyclicity of the underlying graph structure.
This paper follows this direction.
Specifically, the main contributions of the paper are summarized as follows:
\begin{itemize}
   \item We show that an MLP can be seen as an asynchronous message passing GNN applied to the MLP's graph representation.
   \item We propose the \underline{G}raph \underline{N}eural \underline{M}achine (GNM) model, a novel architecture for learning with tabular data which generalizes the MLP architecture and hence, is a family of universal function approximators.
   \item We evaluate the proposed model on several classification and regression datasets where it achieves performance better or comparable to the MLP architecture. %and to other state-of-the-art algorithms for tabular data.
\end{itemize}

%Our contributions are: We show that an MLP can be seen as an asynchronous message passing GNN applied to the MLP's graph representation. From this observation we propose a novel architecture for learning with tabular data which generalizes the MLP architecture, the \underline{G}raph \underline{N}eural \underline{M}achine (GNM) model. Evaluated on several classification and regression datasets, GNMs show better or comparable performance to the MLP architecture.

% The rest of this paper is organized as follows.
% Section~\ref{sec:related_work} summarizes the work related to this paper.
% Section~\ref{sec:preliminaries} introduces some preliminary concepts and summarizes the standard graph neural network model.
% Section~\ref{sec:connection} draws a connection between the MLP architecture and standard GNN models.
% Section~\ref{sec:contribution} presents the proposed model for performing machine learning tasks on tabular data, and shows that it is more general than the standard MLP architecture.
% Section~\ref{sec:experiments} evaluates the proposed architecture on several classification and regression datasets.
% Finally, section~\ref{sec:conclusion} concludes.

\section{Related Work}\label{sec:related_work}
% We begin by briefly reviewing related work. 

\paragraph{Tabular Data Learning with Graphs.} Modelling tabular data with graphs has been an active research direction~\cite{li2024graph}. Tabular data is represented by graphs in mainly three ways: data points as nodes, features as nodes and both as nodes (\ie heterogeneous graphs). Edges are then either constructed based on rules or predicted. We are particularly interested in the feature-as-node scheme~\cite{li2019fi, xie2021bagfn, zhai2023causality} with learned edges, since in our proposed GNM we also represent features as input nodes. 
Xie et al. develop in~\cite{xie2021fives} a predictive model that learns in the meantime multi-scale interactions as weighted connections between input features.
This is achieved by jointly solving the predictive task and the edge prediction task. 
%In our present work, we also train a predictive model while learning an adjacency tensor, yet this work is different from ours in the following aspects: i) We are not limiting our nodes to be only the features, in contrast, we have several types of nodes: input nodes (features), structure nodes and output nodes. The structure nodes in our model can be seen as hidden variables that interact with input features, bringing extra freedom for inference. ii) We learn the graph structure by learning weighted edges and edges having 0 weight are considered not existing. This allows us to establish the link with multi-layer perceptrons and to conclude the important universal approximation property of our model.
Zhai et al. utilize in~\cite{zhai2023causality} the generalized structural equation modeling~\cite{yu2019daggnn} to capture higher order relation between features, including causal relations. A variable graph whose directed edges represent causality relations between two nodes is learned by minimizing the variational lower bound between the true posterior and the variational posterior in a variational Bayes~\cite{kingma2022autoencoding} through the augmented Lagrangian method with $\ell_1$ regularization.

\paragraph{Structural Learning.} Structural learning is an important research area in Bayesian network theory. The objective of this domain is to find directed edges representing causality or dependence among variables. This domain is related to our work since we share the idea of discovering sophisticated interactions between observed or hidden variables. The most important families of structural learning algorithms are~\cite{kitson2023survey}: ($1$) Constraint-based ones, which eliminate edges from dense graphs by conditional independence tests on all pairs of variables like the PC algorithm~\cite{spirtes1991algorithm} and the SGS algorithm~\cite{spirtes2000causation}; and ($2$) Score-based ones, which evaluate candidate structures using a scoring metric, such as the Bayesian Information Criterion or Minimum Description Length~\cite{heckerman1995learning,cooper1999causal}.  

\paragraph{Neural architecture search (NAS).} Neural architecture search consists of automatically discovering the suitable architecture of a neural network. We consider this rapidly evolving research domain related to our work since we share the objective of finding the optimal topological structure of a neural network~\cite{white2023neural}. However, NAS approaches mainly consist of searching optimal architecture as well as other user-defined choices among a tremendous search space using optimized search strategies~\cite{liu2018progressive,christoforidis2021novel,zaidi2021neural,liu2019darts}, in contrast to learning the architecture while training the models as our GNM models do.

\section{Preliminaries}\label{sec:preliminaries}
%We start by fixing our notation, and then we present the family of message passing GNNs.

\subsection{Notation}
Let $\mathbb{N}$ denote the set of natural numbers, \ie $\{1,2,\ldots\}$. 
Then, $[n] = \{1,\ldots,n\} \subset \mathbb{N}$ for $n \geq 1$.
Let also $\oms \cms$ denote a multiset, \ie a generalized concept of a set that allows multiple instances for its elements.
A partition of a set $X$ is a set $P$ of non-empty, non-intersecting subsets of $X$ such that the union of the sets in $P$ is equal to $X$. %, \ie $\bigcup_{A \in P} A = X$. , and the intersection of any two distinct sets in $P$ is empty, \ie for all $A, B \in P$ with $A \neq B$, it holds that $A \cap B = \emptyset$.

Let $G = (V,E)$ be a directed and weighted graph where $V$ is the vertex set and $E$ is the edge set.
%We will denote by $n$ the number of vertices and by $m$ the number of edges, \ie $n = |V|$ and $m = |E|$.
%The adjacency matrix $\mathbf{A} \in \mathbb{R}^{n \times n}$ is a symmetric matrix used to encode edge information in a graph.
%The element of the $i^{\text{th}}$ row and $j^{\text{th}}$ column is equal to the weight of the edge between $v_i$ and $v_j$ if such an edge exists, and $0$ otherwise.
Let $\mathcal{N}^{\leftarrow}(v)$ denote the incoming neighbors of node $v$, \ie the set $\big\{u \colon (u,v) \in E \big\}$.
Let also $\mathcal{N}^{\rightarrow}(v)$ denote the outgoing neighbors of node $v$, \ie the set $\big\{u \colon (v,u) \in E \big\}$.
In the case of directed graphs, every node is associated with an in-degree and an out-degree.
The in-degree $d^{\leftarrow}(v)$ of a node $v$ is equal to the number of incoming edges, \ie $d^{\leftarrow}(v) = |\mathcal{N}^{\leftarrow}(v)|$, while the out-degree $d^{\rightarrow}(v)$ of a node $v$ is equal to the number of outgoing edges, \ie $d^{\rightarrow}(v) = |\mathcal{N}^{\rightarrow}(v)|$.
The nodes of a graph can be annotated with continuous features.
Let $\mathbf{h}_v$ denote the feature of node $v$.
Two graphs $G = (V,E)$ and $G' = (V',E')$ are isomorphic (denoted by $G \cong G'$) if there is a bijective mapping $f : V \rightarrow V'$ such that $\mathbf{h}_v = \mathbf{h}_{f(v)}$ for all $v \in V$ and $(v,u) \in E$ iff $(f(v),f(u)) \in E'$.

\subsection{Graph Neural Networks}
As already discussed, GNNs have recently become the standard approach for dealing with machine learning problems on graphs.
These architectures can naturally model various systems of relations and interactions, including social networks~\cite{kosma2023neural}, particle physics~\cite{shlomi2020graph} and even documents~\cite{nikolentzos2020message}.
A large number of GNN models have been proposed in the past years, and most of these models belong to the family of message passing GNNs~\cite{gilmer2017neural}.
These models employ a message passing (or neighborhood aggregation) procedure to aggregate local
information of nodes.
Suppose each node is annotated with a feature vector, and let $\mathbf{h}_v^{(0)} \in \mathbb{R}^d$ denote node $v$'s initial feature vector.
Then, GNNs iteratively update node representations as follows:
\begin{equation*}
    \begin{split}
        \mathbf{m}_v^{(k)} &= \text{AGGREGATE}^{(k)}  \Big( \bigoms \mathbf{h}_u^{(k-1)} \colon u \in \mathcal{N}(v) \bigcms \Big) \\
        \mathbf{h}_v^{(k)} &= \text{COMBINE}^{(k)}  \Big( \mathbf{h}_v^{(k-1)}, \mathbf{m}_v^{(k)}  \Big)
    \end{split}
    \label{eq:gnn_general}
\end{equation*}
where $\mathcal{N}(v)$ denotes the set of neighbors of node $v$ (in an undirected graphs), and $\text{AGGREGATE}^{(k)}$ is a permutation invariant function of the $k$-th layer that maps the feature vectors of the neighbors of a node $v$ to a single aggregated vector.
This aggregated vector is passed along with the previous representation of $v$ $\big($\ie $\mathbf{h}_v^{(k-1)}\big)$ to the $\text{COMBINE}^{(k)}$ function which combines those two vectors and produces the new representation of $v$.
By defining different $\text{AGGREGATE}^{(k)}$ and $\text{COMBINE}^{(k)}$ functions, we obtain different message passing GNN instances.
In what follows, we use the term GNNs to refer to message passing GNN models.

\section{Connection between MLPs and GNNs}\label{sec:connection}
The MLP architecture has a long history in machine learning and is one of the most-studied learning models.
A multi-layer perceptron is a sequential composition of affine transformations and non-linear activation functions.
It is organized in at least two layers.
Formally, a $K$-layer MLP $f_{\text{MLP}} \colon \mathbb{R}^n \rightarrow \mathbb{R}^c$ is defined as follows:
\begin{equation} \label{eq:MLPupdate}
    f_{\text{MLP}}(\mathbf{x}) = T_K \circ \rho_{K-1} \circ \ldots \circ \rho_1 \circ T_1(\mathbf{x})
\end{equation}
where $T_k \colon \mathbf{x} \mapsto  \mathbf{W}_k \mathbf{x} + \mathbf{b}_k$ is an affine function and $\rho_k$ is a non-linear activation function.

\begin{figure}[t]
    \centering
    \includegraphics[width=.55\textwidth]{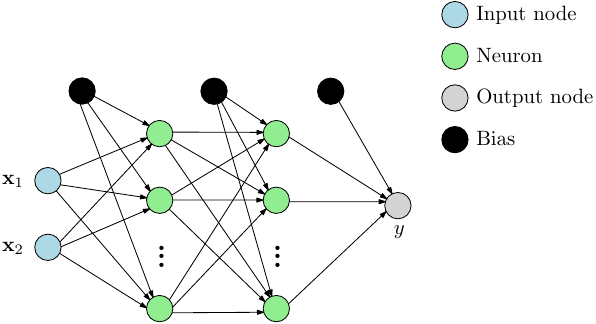}
    \caption{Example of the graph representation of a $3$-layer MLP which can process $2$-dimensional input data.}
    \label{fig:mlp_example}
\end{figure}

In classification or regression problems, we are given a training set $\mathcal{D} = \big\{(\mathbf{x}^{(i)}, \mathbf{y}^{(i)}) \big\}_{i=1}^N$ consisting of $N$ observations along with their class labels or target values, respectively.
Suppose that $\mathbf{x}^{(1)},\ldots,\mathbf{x}^{(N)} \in \mathbb{R}^m$ and $\mathbf{y}^{(1)},\ldots,\mathbf{y}^{(N)} \in \mathbb{R}^c$.
A $K$-layer MLP for such a classification or regression problem can be represented as a directed acyclic graph $G_{\text{MLP}}=(V_{\text{MLP}},E_{\text{MLP}})$ where nodes represent the input features, the hidden neurons, the biases, and the output neurons.
Therefore, the set of nodes $V_{\text{MLP}}$ can be partitioned into the following pairwise disjoint sets that consist of the aforementioned four types of nodes: $\mathcal{X}, \mathcal{H}, \mathcal{B}$ and $\mathcal{Y}$, each set has the cardinality being  $|\mathcal{X}| = m$, $|\mathcal{Y}| = c$, $|\mathcal{B}| = K$, and $|\mathcal{H}|$ depends on the values of the MLP's hyperparameters set by the user.
The number of nodes of $G_\text{MLP}$ is equal to $m + n_1 + \ldots + n_{K-1} + K + c$ where $n_1, \ldots, n_K$ denote the number of neurons of the $1$-st, $\ldots$, $(K-1)$-th hidden layer of the model.
The $G_{\text{MLP}}$ graph representation of a $3$-layer MLP which can process $2$-dimensional input data is illustrated in Figure~\ref{fig:mlp_example}.

We next establish a connection between MLPs and GNNs. 
There is very recent work discussing how neural networks can be represented as graphs, and how they can be incorporated as inputs to various GNN models~\cite{zhang2023neural}. 
Unlike this work, our focus is 
not on applying different GNNs to the graph representation of MLPs, but instead
on moving away from the standard directed acyclic graph representation, and as a result, generalizing MLPs.
We will next show that training an MLP in a classification or regression task is equivalent to training an asynchronous GNN in a node classification or node regression problem, respectively.

Given an input sample $\mathbf{x} \in \mathbb{R}^m$, we now discuss how we can use a GNN to compute the output $\hat{\mathbf{y}} \in \mathbb{R}^c$.
The nodes of the graph $G_{\text{MLP}}$ are annotated with scalar features as follows.
The $m$ nodes that belong to set $\mathcal{X}$ are annotated with the values of the features of the input sample $\mathbf{x}$ (\ie $\mathbf{x}_1, \ldots,\mathbf{x}_m$).
The $K$ nodes that belong to set $\mathcal{B}$ are annotated with a value equal to $1$.
The rest of the nodes are annotated with a value equal to $0$.
Note that each input sample $\mathbf{x}^{(i)}$ is associated with a graph $G_{\text{MLP}}^{(i)}$.
Thus, the graphs $G_{\text{MLP}}^{(i)}$ and $G_{\text{MLP}}^{(j)}$ 
%associated with two input samples $\mathbf{x}^{(i)}$ and $\mathbf{x}^{(j)}$, respectively, 
are isomorphic to each other, \ie $G_{\text{MLP}}^{(i)} \cong G_{\text{MLP}}^{(j)}$ iff $\mathbf{x}^{(i)} = \mathbf{x}^{(j)}$.

\begin{figure}[t]
    \centering
    \includegraphics[width=\textwidth]{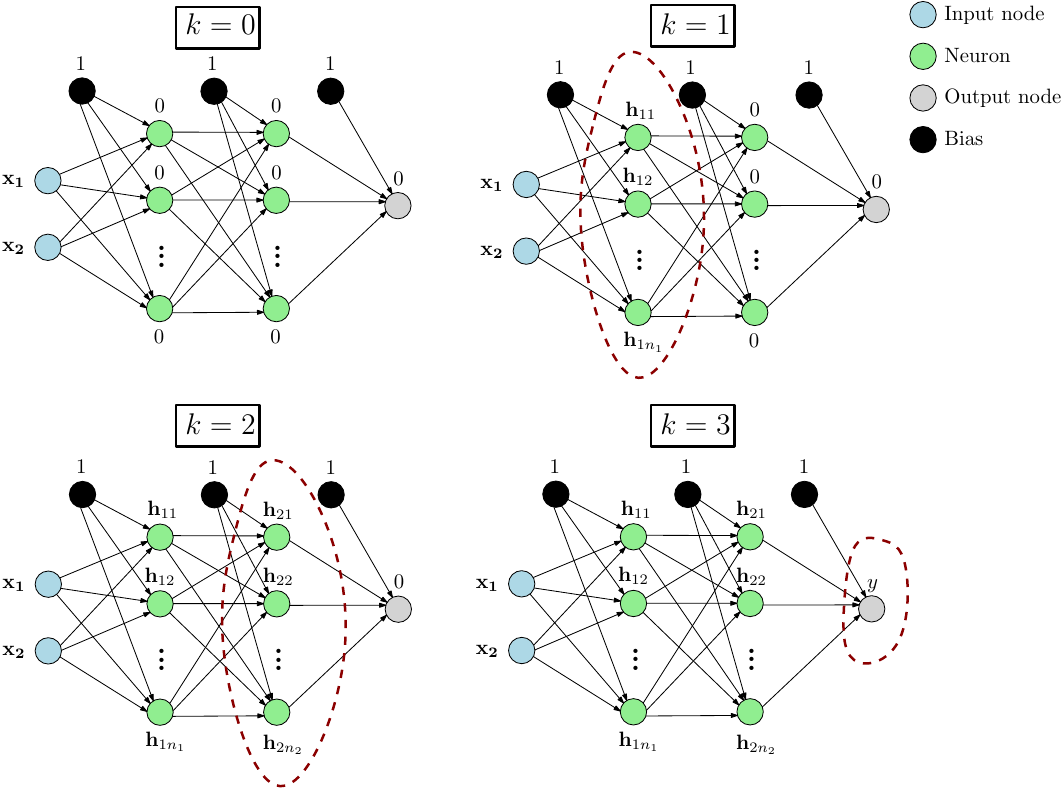}
    \caption{An illustration of the computation of a $3$-layer MLP as an asynchronous message passing procedure.}
    \label{fig:mlp_prop}
\end{figure}

We use the notation GNN\textsubscript{MLP} to denote the GNN model that is defined in Equation (\ref{eq:AGG_MLP}).
The $\text{AGGREGATE}$ function of the GNN\textsubscript{MLP} model is defined as follows:
\begin{equation}\label{eq:AGG_MLP}
    h_v^{(k)} = f \bigg( \sum_{u \in \mathcal{N}^{\leftarrow}(v)} \alpha_{uv} \, h_u^{(k-1)} \bigg) 
\end{equation}
where $f$ is an activation function and $\alpha_{uv} \in \mathbb{R}$ is a trainable parameter for the pair of nodes $u$ and $v$.
Therefore, the trainable parameters of the model correspond to the set $\theta = \big\{ \alpha_{uv} \colon (u,v) \in E_{\text{MLP}} \big\}$.
The set $\theta$ can be thought of as attention coefficients that allow each node to attend to its neighbors.
Note that the GNN\textsubscript{MLP} model does not utilize any $\text{COMBINE}$ function.
Note also that $h_v^{(0)} \in \mathbb{R}$ for all $v \in V_\text{MLP}$.
Therefore, we also have that $h_v^{(t)} \in \mathbb{R}$ for all $v \in V$ and for all $t \in [K]$.
To compute the final node representations, an asynchronous message-passing scheme is employed, where nodes are updated based on their position in the network.
The $n_1$ nodes representing the first hidden layer are updated first.
Then, the $n_2$ nodes of the second hidden layer are updated, and so on.
Once the representations of the $n_{K-1}$ nodes of the $(K-1)$-th layer have been updated, we finally compute the new representations of the output nodes of set $\mathcal{Y}$.
This asynchronous message passing procedure is illustrated in Figure~\ref{fig:mlp_prop}.

A classification or a regression task can be thus formulated as a node classification or node regression task, where the final features of the output nodes are compared against the ground-truth labels or targets to compute the error.
Formally, in a regression task, we can compute the loss as follows:
\begin{equation}\label{eq:lossRegress}
    \mathcal{L} = \frac{1}{N} \sum_{i=1}^N \sum_{j=1}^c \Big( \mathbf{y}_j^{(i)} - h_{g(j)}^{(K)} \Big)^2
\end{equation}
where $g : [c] \rightarrow \mathcal{Y}$ is a function that maps classes/targets to the corresponding nodes of $G_{\text{MLP}}$.
Likewise, in a multi-class classification problem, we first apply the softmax function to the outputs:
\begin{equation*}
    \bar{h}_{g(i)} = \frac{e^{h_{g(i)}^{(K)}}}{\sum_{j=1}^c e^{h_{g(j)}^{(K)}}}
\end{equation*}
And then, we can use the negative log likelihood of the multinomially distributed correct labels, i.e., the cross entropy, as training loss:
\begin{equation} \label{eq:lossClassif}
    \mathcal{L} = - \sum_{i=1}^N \sum_{j=1}^c \mathbf{y}_j^{(i)} \log \bar{h}_{g(j)}^{(K)}
\end{equation}

The now following Theorem \ref{thm:anMlpIsGNM} formalizes the relationship between MLPs and GNNs.

\begin{theorem}\label{thm:anMlpIsGNM}
    Let $G_{\text{MLP}}$ be the graph associated with a $K$-layer MLP.
    Then, the $K$-layer MLP is equivalent\footnote{A model is characterized by its update mechanism. Two models are equivalent if their update mechanisms are identical.} to a GNN\textsubscript{MLP} applied to $G_{\text{MLP}}$ with the aggregation function (\ref{eq:AGG_MLP}) where $f$ is the activation function of the MLP.
\end{theorem}

Hence, MLPs can be identically reproduced by GNNs with the aggregate function in Equation (\ref{eq:AGG_MLP}). The proof of Theorem \ref{thm:anMlpIsGNM}, and all subsequent theoretical claims, can be found in Appendix~\ref{app:Proofs}. 

\section{Graph Neural Machines}\label{sec:contribution}
In the previous section, we showed that an MLP can be seen as a GNN that operates on a directed acyclic graph.
In this section, we propose a new machine learning model for tabular data, the so-called \underline{G}raph \underline{N}eural \underline{M}achine (GNM), which generalizes the MLP architecture by (1) replacing the directed acyclic graph $G_{\text{MLP}}$ with a nearly complete directed graph which also contains directed cycles; and (2) dropping the asynchronous message passing scheme and employing a standard synchronous scheme.

\begin{figure}[t]
    \centering
    \includegraphics[width=0.55\columnwidth]{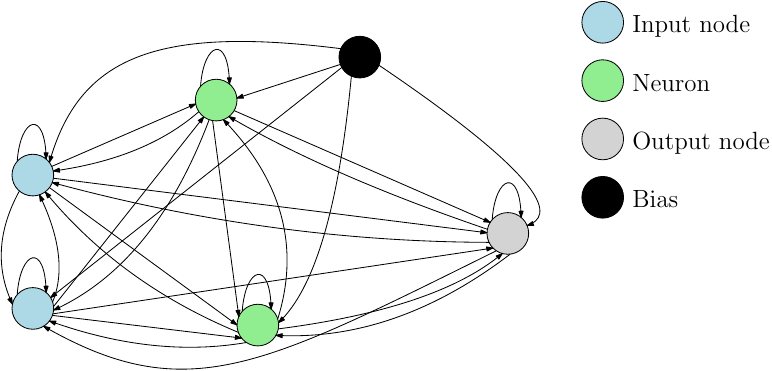}
    \caption{An example of the graph representation employed by the proposed GNM model.}
    \label{fig:gnm_graph}
\end{figure}

We first replace graph $G_{\text{MLP}}$ with graph $G_{\text{GNM}}$.
The graph $G_{\text{GNM}}=(V_{\text{GNM}}, E_{\text{GNM}})$ is a directed (but not acyclic) graph where nodes represent the input features, the hidden neurons, the bias, and the output neurons.
Therefore, similar to $G_{\text{MLP}}$, the set of nodes $V_{\text{GNM}}$ can also be partitioned into the following pairwise disjoint sets which consist of the aforementioned four types of nodes: $\mathcal{X}, \mathcal{H}, \mathcal{B}$ and $\mathcal{Y}$.
The subgraph of $G_{\text{GNM}}$ induced by the set $\mathcal{X} \cup \mathcal{H} \cup \mathcal{Y}$ is a complete directed graph, \ie a directed graph in which every pair of distinct nodes is connected by a pair of unique edges (one in each direction).
This subgraph also contains self-loops, \ie for each node, there exists an edge that connects the node to itself.
The set $\mathcal{B}$ contains only a single node.
There are directed edges from this node to all the other nodes of the graph.
An example of a graph $G_{\text{GNM}}$ which consists of two input nodes, one bias node, two neurons and an output node is illustrated in Figure~\ref{fig:gnm_graph}.

Given an input sample $\mathbf{x} \in \mathbb{R}^m$, the nodes of the $G_{\text{GNM}}$ graph are annotated with features as follows.
The $m$ nodes that belong to set $\mathcal{X}$ are annotated with the values of the features of the input sample $\mathbf{x} \in \mathbb{R}^m$ (\ie $\mathbf{x}_1, \ldots, \mathbf{x}_m$).
The single node that belongs to set $\mathcal{B}$ is annotated with a value equal to $1$.
The rest of the nodes are annotated with a value equal to $0$.

We next describe how the output is computed given an input sample $\mathbf{x}$.
More specifically, the GNM model consists of a series of message passing layers.
Suppose there are $K$ such layers in total.
All these layers employ the same $\text{AGGREGATE}^{(k)}$ function.
Specifically, in the $k$-th layer, the feature $h_v^{(k)}$ of a node $v$ (for all nodes except the bias node) is updated as follows:
\begin{equation}
    h_v^{(k)} = f \bigg( \sum_{u \in \mathcal{N}^{\leftarrow}(v)} \alpha_{uv}^{(k)} \, h_u^{(k-1)} \bigg)
    \label{eq:gnm_update}
\end{equation}
Note that different weights $\alpha_{uv}^{(k)}$ are learned per node pair per layer.
Therefore, we can have $\alpha_{uv}^{(i)} \neq \alpha_{uv}^{(j)}$ for $i,j \in [K]$ and $i \neq j$.
Furthermore, $\forall v \in V \setminus \mathcal{B}$, the update rule of Equation~\eqref{eq:gnm_update} is equivalent to the following:
\begin{equation*}
    h_v^{(k)} = f \bigg( \sum_{u \in V} \alpha_{uv}^{(k)} \, h_u^{(k-1)} \bigg)
\end{equation*}
since the in-degree of all these nodes is equal to the number of nodes of the graph.
With regards to the bias node, there are no incoming edges to that node, and in fact, we choose not to update the feature of that node.
In other words, $h_{v_b}^{(k)} = 1$ for any $k \in [K] \cup \{0\}$, where $v_b$ denotes the bias node.
The update function in Equation (\ref{eq:gnm_update}) can be also written as:
\begin{equation*}
    \mathbf{h}^{(k)} = f \Big( \mathbf{A}^{(k)} \, \mathbf{h}^{(k-1)} \Big)
\end{equation*}
where $\mathbf{h}^{(0)},\ldots,\mathbf{h}^{(K)} \in \mathbb{R}^n,$ $\mathbf{A}^{(1)},\ldots,\mathbf{A}^{(K)} \in \mathbb{R}^{n \times n}$ and $n$ denotes the number of nodes of $G_{\text{GNM}}$.
The $i$-th element of $\mathbf{h}^{(k)}$ stores the feature of the $i$-th node at layer $k$, while the element of the $i$-th row and $j$-th column of $\mathbf{A}^{(k)}$ is equal to the trainable weight of the edge from the $j$-th to the $i$-th node if such an edge exists, and $0$ otherwise (or $1$, in case $i=j$ or the bias node is the graph's $i$-th node).
Without loss of generality, suppose the bias node is the last node of the graph (\ie $n$-th node).
Then, matrix $\mathbf{A}^{(k)}$ would be equal to:
\[
\mathbf{A}^{(k)} = 
\begin{bmatrix}
    \alpha_{11}^{(k)} & \alpha_{21}^{(k)} & \cdots & \alpha_{n1}^{(k)} \\
    \alpha_{12}^{(k)} & \alpha_{22}^{(k)} & \cdots & \alpha_{n2}^{(k)}\\
    \vdots & \vdots & \ddots & \vdots \\
    0 & 0 & \cdots & 1
\end{bmatrix}
\]   

The output of the model for a given input vector $\mathbf{x}$ corresponds to the final representations of the output nodes, \ie $h_v^{(K)}$ for $v \in \mathcal{Y}$.
The features of these nodes 
%can be compared against the ground-truth values 
are used to compute the loss and we train the model as discussed in Section \ref{sec:connection}.

\paragraph{Expressive power.}
We next provide some theoretical results about the expressive power of the proposed GNM model.
We first show that when the number of hidden neurons is greater than $1$, a single GNM model can simulate several different MLPs.
From that, we derive one of our most important contributions of the present work, namely that GNM is a family of universal approximators. We further give hints about the potentially stronger expressive power of GNM than MLP under a similar degree of freedom by showing that the GNM model class strictly includes MLP class.

% \begin{theorem}
%     GNM can simulate any $K$-layer MLP with the same number of neurons distributed across the different layers.
% \end{theorem} 
% @Giannis, I would like to reformulate the theorem as below:
\begin{theorem} \label{thm:anyMLPIsGNM}
    Let $\text{MLP}_{N,K}$ and $\text{GNM}_{N,K}$ denote, respectively, the set of all MLPs having $N$ non-bias neurons arranged into $K$ layers and the set of all $K$-layer GNMs having $N$ non-bias nodes. Then, $\forall M_{\text{MLP}} \in \text{MLP}_{N,K},\ \exists M_{\text{GNM}} \in \text{GNM}_{N,K}$ such that $M_{\text{GNM}}$ represents $M_{\text{MLP}}$, \ie $M_{\text{GNM}}(\mathbf{x}) = M_{\text{MLP}}(\mathbf{x})\ \forall \mathbf{x} \in \mathbb{R}^m$. 

\end{theorem}

Therefore, we show that by allowing a dense graph to have different sets of edge weights for each update step, MLPs can be seen as a special case of GNM. This leads to the following corollary.
\begin{corollary}\label{cor:univ_approx}
    GNM is a family of universal approximators of continuous functions defined on compact sets.
\end{corollary}

Note that even though the universal approximation ability of the GNM family follows by showing that MLPs are a subset of GNMs, this does not imply that a GNM necessarily approximates a target function by representing MLPs. 

In fact, more interestingly, the parameter space of GNM strictly includes any MLP adjacency matrices representing arrangements of $N$ (non-bias) nodes into $K$ layers, i.e., MLPs are actually a proper subset of GNM. Indeed, a frame $\mathbf{A}$ of an adjacency tensor $\mathcal{A}$ having non-zero terms on its diagonal blocks is a valid GNM. To state this in the MLP context, non-zero diagonal blocks in an adjacency matrix $\mathbf{A}$ correspond to connections within a hidden layer, making the structure no longer a multi-partite graph. Another example of the strict inclusion would be a case where a neuron can be updated at different layers in a GNM, whereas in an MLP, each neuron gets updated once and only once during a forward propagation. These examples indicate that GNMs have potentially stronger expressive power than MLPs.

\paragraph{Running time.}
%In each layer $k$ of the model, matrix $\mathbf{A}^{(k)}$ needs to be multiplied by vector $\mathbf{h}^{(k-1)}$.
Each update $k$ of the model requires a multiplication of matrix $\mathbf{A}^{(k)}$ by vector $\mathbf{h}^{(k-1)}$.
The complexity of this operation is $\mathcal{O}(n^2)$ where $n$ is the number of nodes of $G_{\text{GNM}}$.
Therefore, computing the output of the model requires time $\mathcal{O}(K n^2)$.
However, we can add some regularization term to the model's loss function such that some parameters are set to $0$ (as shown in Section~\ref{sec:regularization}).
At inference time, we can store matrix $\mathbf{A}^{(k)}$ as a sparse matrix, and then, computing the output requires time $\mathcal{O}(\mathfrak{m}^{(1)}+\ldots+\mathfrak{m}^{(K)})$ where $\mathfrak{m}^{(i)}$ denotes the number of non zero elements of matrix $\mathbf{A}^{(i)}$.
We empirically measure the running time of GNM and compare it against that of MLP in Appendix~\ref{sec:complexity}.

\paragraph{Limitations.}
The relatively large parameter space can imply higher flexibility of the GNM model, but it comes with a cost of high risk of overfitting% and non-identifiability.
One way to mitigate this is to apply regularization.  
Regularizing MLPs in order to gain computational efficiency and enhance generalization has been extensively studied in the past. One of the most applied techniques is Dropout \cite{srivastava2014dropout}; other works focus on well-designed loss functions using $\ell_i$ penalties with $i \in \{1,2\}$ (\ie $\ell_1$ penalty: LASSO \cite{koneru2021smoothed,ochiai2017automatic}, $\ell_2$ penalty: Ridge \cite{warton2008penalized} and combined: Elastic penalties \cite{zou2005regularization}). Augmenting the loss functions from (\ref{eq:lossRegress}) and (\ref{eq:lossClassif}) to $\mathcal{L} + \lambda \Vert \theta \Vert_i^i$ 
%where $\theta$ contains all learnable parameters 
encourages the optimization algorithm to shrink parameter values close or equal to $0$. To ensure sparsity, networks are often further pruned, \ie edge weights having absolute value smaller than a threshold are set to $0$ \cite{han2015learning}. More recent works introduced $\ell_0$ regularization that circumvents the parameter value shrinkage problem in $\ell_{1,2}$ penalties, \ie parameter absolute values are small instead of $0$, and directly penalizes non zero terms \cite{louizos2018learning}. 
%To the best of our knowledge, these efforts have never been applied to dense graphs having node features updated by learned weights. 
Our hope is that with well-designed regularization, the GNM model can learn meaningful computational graph structures that are not necessarily acyclic multi-partite. We evaluate GNM's robustness to overfitting and carry out experiments on synthetic datasets in Section~\ref{sec:OverfittingResults}. %and provide evidence for this hypothesis 

Though our proposed GNM is seemingly less computationally efficient than even dense MLPs, we see some advantages therein. Firstly, universal approximation and expressive power studies about deep neural networks suggest that their ability to fit a target function can be sensible to the structure design \cite{raghu2017expressive}; however a GNM, in theory, will find an optimal structure during training. Secondly, to further encourage a network that has both sparse and effective connections we can regularize GNMs as inspired by previous works for MLPs. 
%More importantly, we achieve such a model by pruning parameters while training, not prior to training by random sampling, nor posterior by eliminating weights with absolute value smaller than some threshold, the resulting sparse structure will be a data-driven decision, rather than a random choice which could potentially harm the universal approximation ability of an MLP. 
Thirdly, we no longer forbid connections within a layer as in MLPs, as we do not have the notion of layers in our graph structure. This could bring higher expressive power to the model.

\section{Experimental Evaluation}\label{sec:experiments}
We now empirically compare the MLP and GNM models. 

\subsection{Datasets}
We evaluated the proposed model on $15$ classification datasets: Noisy Moons, Abalone, Adult, Car, Connect4, Phishing, Wireless, Yeast, PageBlocks, DryBean, Letter Recognition, Electrical Grid Stability, Susy, Isolet and Waveform.
The Noisy Moons dataset is a synthetic dataset, while the rest of the datasets are publicly available in the UCI repository~\cite{asuncion2007uci}.
We also evaluated the GNM model on $6$ regression datasets, namely Wine, Auto MPG, Bike Sharing, Diabetes, California and Year.
The Diabetes and California datasets are accessed through scikit-learn~\cite{pedregosa2011scikit}, while the rest of the datasets are obtained from the UCI repository.
%For all datasets, we transform categorical variables to one-hot vectors.
We used one-hot encoding for categorical variables in all datasets.
More details about the datasets are given in Appendix~\ref{sec:datasets}.

\subsection{Experimental Setup}
To evaluate the different models, we performed $10$-fold cross-validation.
Within each fold, $10\%$ of the training samples were used as validation set.
All models were tested on the same splits.

With regards to the hyperparameters of the models, for GNM, we chose the number of nodes (including input and output nodes) from $\{50,100,200,300\}$.
For MLP, we chose the number of hidden units from $\{32,64,128,256\}$.
On the Isolet dataset where the number of features $m$ is equal to $617$, we chose the number of nodes of GNM and the number of hidden units of the MLP from $\{800,1000\}$ and $\{512,1024\}$, respectively.
For both models, we also tuned the following two hyperparameters for each dataset: ($1$) the number of layers $\in \{2,3,4\}$; and ($2$) the dropout ratio $\in \{0, 0.2\}$.
We applied the ReLU activation function to the representations of the neurons that emerge in all but the last layer.
We set the batch size to $64$.
We trained each model for $300$ epochs.
We used the Adam optimizer and chose the learning rate from $\{0.01,0.001\}$.
We used the binary or categorical cross-entropy loss in classification tasks and the mean squared error (MSE) loss in regression tasks.
Within each fold, we chose the model that achieves the lowest loss on the validation set and evaluated the model on the test set.
For a fair comparison, we used a model parameter budget of about $500$k parameters ($5$m parameters for the Isolet dataset).

The models are implemented in Python using PyTorch~\cite{paszke2019pytorch}.
The code is available at \url{https://github.com/giannisnik/gnm}.
All experiments were conducted on a single machine with $16$ cores, $128$ GBs of RAM, and 2 NVIDIA RTX A6000 GPUs with $48$ GBs of memory.

\begin{table}[t]
    \centering
    \caption{Average classification accuracy and macro F1-score ($\pm$ standard deviation) of the MLP and GNM architectures on the $15$ classification datasets.}
    \vspace{.1cm}
    \footnotesize
    \label{tab:classification_results}
    \renewcommand{\arraystretch}{1.2}
    \begin{tabular}{l|cc|cc}
    \toprule
    & \multicolumn{2}{c}{MLP} & \multicolumn{2}{|c}{GNM} \\ \cline{2-5}
    & Acc & F1 & Acc & F1  \\
    \midrule
    Noisy Moons & 99.90 $\pm$ 0.30 & 99.89 $\pm$ 0.30 & \textbf{100.0} $\pm$ 0.00 & \textbf{100.0} $\pm$ 0.00 \\
    Abalone & \textbf{27.69} $\pm$ 1.98 & \textbf{14.93} $\pm$ 2.30 & 27.64 $\pm$ 1.93 & 14.56 $\pm$ 1.98 \\
    Adult & 84.55 $\pm$ 0.37 & 76.20 $\pm$ 0.96 & \textbf{85.06} $\pm$ 0.25 & \textbf{77.94} $\pm$ 0.58 \\
    Car & 99.07 $\pm$ 0.78 & 97.31 $\pm$ 1.77 & \textbf{99.36} $\pm$ 0.48 & \textbf{98.23} $\pm$ 2.18 \\
    Connect4 & \textbf{84.58} $\pm$ 0.74 & 67.44 $\pm$ 2.07 & 84.32 $\pm$ 0.54 & \textbf{68.09} $\pm$ 1.55 \\
    Phishing & \textbf{96.91} $\pm$ 0.67 & \textbf{96.86} $\pm$ 0.68 & 96.79 $\pm$ 0.51 & 96.74 $\pm$ 0.53 \\
    Wireless & 97.60 $\pm$ 1.06 & 97.56 $\pm$ 1.06 & \textbf{97.70} $\pm$ 1.22 & \textbf{97.65} $\pm$ 1.22 \\
    Yeast & 58.82 $\pm$ 5.02 & 48.99 $\pm$ 8.92 & \textbf{59.56} $\pm$ 4.55 & \textbf{51.63} $\pm$ 6.16 \\
    PageBlocks & 95.88 $\pm$ 1.06 & 64.33 $\pm$ 8.35 & \textbf{96.27} $\pm$ 0.78 & \textbf{65.15} $\pm$ 4.65 \\
    DryBean & 86.26 $\pm$ 1.09 & 86.21 $\pm$ 1.62 & \textbf{89.14} $\pm$ 0.93 & \textbf{89.80} $\pm$ 0.87 \\
    Letter Recognition & \textbf{93.50} $\pm$ 0.81 & \textbf{93.45} $\pm$ 0.80 & 92.84 $\pm$ 0.60 & 92.80 $\pm$ 0.65 \\
    Electrical Grid Stability & 99.59 $\pm$ 0.15 & 99.55 $\pm$ 0.16 & \textbf{99.79} $\pm$ 0.13 & \textbf{99.77} $\pm$ 0.14 \\
    Susy & 78.65 $\pm$ 0.71 & 77.60 $\pm$ 0.90 & \textbf{78.88} $\pm$ 0.57 & \textbf{77.85} $\pm$ 0.67 \\
    Isolet & \textbf{94.99} $\pm$ 1.14 & \textbf{94.90} $\pm$ 1.17 & 94.85 $\pm$ 0.66 & 94.78 $\pm$ 0.73 \\
    Waveform & \textbf{85.10} $\pm$ 2.06 & \textbf{85.00} $\pm$ 2.05 & 85.06 $\pm$ 1.62 & \textbf{85.00} $\pm$ 1.62 \\
    \bottomrule
    \end{tabular}
\end{table}

\subsection{Experimental Results}
Table~\ref{tab:classification_results} illustrates the obtained results on the different classification datasets.
We report the average accuracy and average macro F1-score and corresponding standard deviations on the test set across the $10$ folds within the cross-validation.
Overall, we observe that, with equal parameter budgets, the proposed GNM model outperforms the MLP architecture.
More specifically, in terms of accuracy, GNM is the best performing model on $9$ out of the $15$ datasets, while in terms of F1-score, it outperforms the MLP on $10$ out of the $15$ classification datasets.
As expected, the difference in performance between the two models is not very large, but in some cases, GNM provides significant improvements.
For instance, on the DryBean dataset, GNM provides an absolute increase of $2.88\%$ in accuracy over the MLP model.

Table~\ref{tab:regression_results} illustrates the obtained results on the $6$ regression datasets.
We report the average MSE and average R2 score and corresponding standard deviations on the test set across the $10$ folds within the cross-validation.
In the regression task, the proposed GNM model significantly outperforms the MLP architecture.
GNM performs better in terms of MSE on $5$ out of the $6$ considered datasets and is outperformed by the MLP only on the Diabetes dataset.
Overall, our results indicate that the proposed model is empirically stronger than the MLP. 
Improvements are more visible in regression tasks than classification tasks.
We believe that this is because classification tasks allow larger margins for different separation boundary functions.
Therefore, a model that approximately finds the boundaries already achieves high levels of performance.

\begin{table}[t]
    \centering
    \caption{Average mean squared error and R2 score ($\pm$ standard deviation) of the MLP and GNM architectures on the $6$ regression datasets.}
    \vspace{.1cm}
    \footnotesize
    \label{tab:regression_results}
    \renewcommand{\arraystretch}{1.2}
    \begin{tabular}{l|cc|cc}
    \toprule
    & \multicolumn{2}{c}{MLP} & \multicolumn{2}{|c}{GNM} \\ \cline{2-5}
    & MSE & R2 & MSE & R2  \\
    \midrule
    Wine & \phantom{00}0.572 $\pm$ 0.102\phantom{00} & \phantom{-}0.271 $\pm$ 0.109 & \textbf{\phantom{00}0.540} $\pm$ 0.031\phantom{00} & \textbf{\phantom{-}0.310} $\pm$ 0.030 \\
    Auto MPG & \phantom{0}12.710 $\pm$ 3.743\phantom{00} & \phantom{-}0.783 $\pm$ 0.078 & \textbf{\phantom{0}10.124} $\pm$ 2.947\phantom{00} & \textbf{\phantom{-}0.828} $\pm$ 0.060 \\
    Bike sharing & 0.00044 $\pm$ 0.00031 & \textbf{\phantom{-}0.999} $\pm$ 0.000 & \textbf{0.00039} $\pm$ 0.00002 & \textbf{\phantom{-}0.999} $\pm$ 0.000 \\
    Diabetes & \textbf{\phantom{0}3173.8} $\pm$ 618.1\phantom{00} & \textbf{\phantom{-}0.445} $\pm$ 0.129 & \phantom{0}3187.9 $\pm$ 618.1\phantom{00} & \phantom{-}0.443 $\pm$ 0.127 \\
    California & \phantom{00}0.460 $\pm$ 0.036\phantom{00} & \phantom{-}0.653 $\pm$ 0.023 & \textbf{\phantom{00}0.436} $\pm$ 0.033\phantom{00} & \textbf{\phantom{-}0.671} $\pm$ 0.024 \\
    Year & \phantom{00}989.7 $\pm$ 177.7\phantom{00} & -7.282 $\pm$ 1.481 & \textbf{\phantom{00}240.0} $\pm$ 161.8\phantom{00} & \textbf{-1.013} $\pm$ 1.363 \\
    \bottomrule
    \end{tabular}
\end{table}

\begin{figure}[t]
    \centering
    \includegraphics[width=.8\textwidth,trim={0 0.2cm 0 0}]{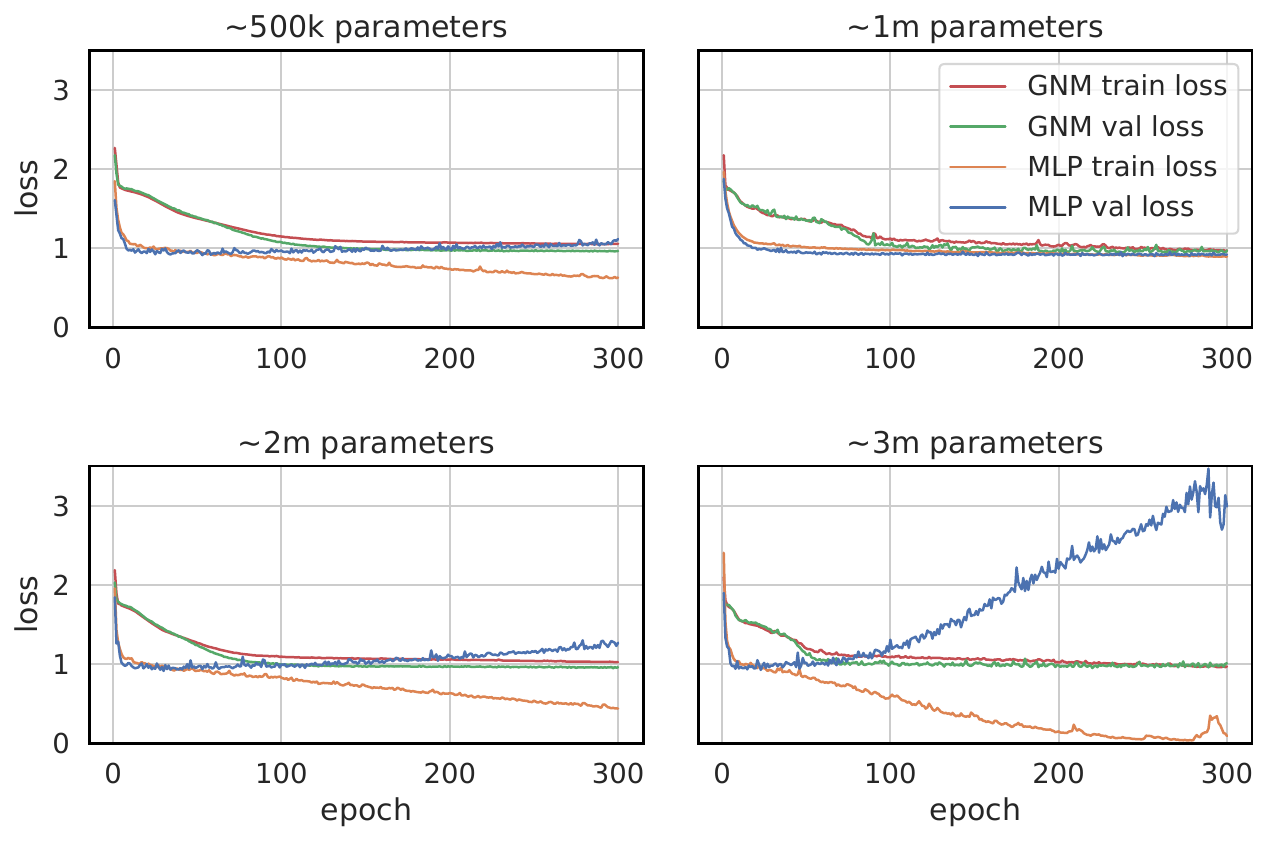}
    \caption{Training curves of the GNM and MLP models on the Yeast dataset. Four instances of each model are trained consisting of approximately $500$k, $1$m, $2$m and $3$m parameters.}
    \label{fig:overfitting}
\end{figure}

\subsection{Robustness to Overfitting} \label{sec:OverfittingResults}
We next investigate whether the proposed model and the MLP become more prone to overfitting as the number of parameters increases.
More specifically, we trained $4$ instances of each model on one split of the Yeast dataset.
We set the hyperparameters to values such that the $4$ instances consist of approximately $500$k, $1$m, $2$m, and $3$m parameters.
Figure~\ref{fig:overfitting} illustrates the loss on the training and validation sets achieved by the two models across the $300$ epochs.
We observe that the MLP suffers significantly from overfitting when the number of parameters is large (\ie $2$m and $3$m).
Even when the number of parameters is around $500$k, there is some difference between the loss on the training set and that on the validation set.
On the other hand, the proposed GNM model seems to be less prone to overfitting.
For all number of parameters, the loss on the two sets is close.
Results from $4$ more datasets are provided in Appendix~\ref{sec:overfitting}.
It generally appears that GNM does not suffer from overfitting as much as the MLP model.

\subsection{Learning Sparse GNM Models}\label{sec:regularization}
We next study the potential of learning sparse GNM models. 
To achieve sparsity, we added an $\ell_1$ regularization term to the loss function, \ie we added the term $\lambda \sum_{k=1}^K \sum_{i=1}^{n-1} \sum_{j=1}^n |\mathbf{A}_{ij}^{(k)}|$ where $\lambda$ is a hyperparameter that determines the contribution of the penalty term.
This regularization scheme is typically useful for feature selection since it forces the model to shrink the coefficients of the less important features towards zero.
In this experiment, we constructed a binary classification problem (a variant of the classic XOR problem) where $200$ data points are evenly distributed into $2$ classes.
Each data point is a $2$-dimensional vector.
%Half of the data points of the first class are sampled from a Gaussian distribution with mean $\mu = [1,1]$, while the rest of the data points are sampled from a Gaussian distribution with mean $\mu = [-1,-1]$.
%With regards to the second class, half of its data points are sampled from a Gaussian distribution with mean $\mu = [1,-1]$, while the rest of the data points are sampled from a Gaussian distribution with mean $\mu = [-1,1]$
%All four distributions share identical covariance matrices.
%Each covariance matrix is a diagonal matrix with a value equal to $0.3$ on the main diagonal.
The first class contains two sets of $50$ i.i.d. samples from two Gaussian distributions $\mathcal{N}(\boldsymbol{\mu}_1, \Sigma)$ and $\mathcal{N}(\boldsymbol{\mu}_2, \Sigma)$ respectively, with $\boldsymbol{\mu}_1 = [1,1]$ and $\boldsymbol{\mu}_2 = [-1,-1]$.
The second class contains also two sets of $50$ i.i.d. samples from two Gaussian distributions $\mathcal{N}(\boldsymbol{\mu}_3, \Sigma)$ and $\mathcal{N}(\boldsymbol{\mu}_4, \Sigma)$ respectively, with $\boldsymbol{\mu}_3 = [-1,1]$ and $\boldsymbol{\mu}_4 = [1,-1]$.
The covariance matrices of all four distributions was set to $0.3\mathbf{I}$.
We split the dataset into a training and a test set using a $90-10$ split.
The number of nodes of the model was set to $50$ and the number of layers was set to $2$.
We added the $\ell_1$ regularization term to the loss function.
To make predictions for the test samples, we set all elements of $\mathbf{A}^{(1)}$ and $\mathbf{A}^{(2)}$ whose absolute value is smaller than $10^{-3}$ to zero.
The accuracy of the GNM model on the test set turned out to be $100\%$.
We also retrieved the positions of the nonzero elements of matrices $\mathbf{A}^{(1)}$ and $\mathbf{A}^{(2)}$.
For $\mathbf{A}^{(1)}$, we obtained the following positions: 
$(0,  3)$, $(0, 15)$, $(0, 16)$, $(0, 39)$, $(1, 3)$, $(1,15)$, $(1, 16)$, $(1, 39)$, $(48, 3)$, $(48, 39)$, while for $\mathbf{A}^{(2)}$, the elements in the following positions were different from zero: $(3, 49)$, $(15, 49)$, $(16, 49)$, $(39, 49)$.
Note that rows $0$ and $1$ correspond to the two input nodes, row $49$ corresponds to the output node, while row $48$ corresponds to the bias node.
Therefore, the emerging GNM model is equivalent to a $2$-layer MLP that contains $4$ hidden units and no bias in the second layer. 
\section{Conclusion}\label{sec:conclusion}
In this work, we presented the GNM model, a universal approximator that generalizes standard MLPs by allowing a wider range of graph structures as the computational graph. This opens up the possibility of exploring how the topology of a graph can impact the efficiency of learning and the expressive power of a model. The freedom of automatically learning graph structure being left to the model allows to save considerable effort for hyper-parameter tuning often required in MLP architecture designs. We have discussed the limitations of the model brought by its large number of learnable parameters.
We consider regularization as a promising solution and in fact, we empirically show that by applying regularization and pruning, a large number of computational nodes can be dropped.
Besides the theoretical analysis, we also empirically evaluate the proposed model in classification and regression datasets.
Our results indicate that in most cases, the GNM model achieves better or comparable performance with the MLP architecture.
%However, to our surprise, we experimentally show that with comparable number of trainable parameters, GNM is less prone to over-fit than MLP, this could suggest that our model truly learns rather than memorises.

\bibliography{biblio}
\bibliographystyle{plain}

\newpage
\appendix

\section{Proofs}\label{app:Proofs}

We now provide the proofs of the theoretical claims made in the main paper. 
\subsection{Proof of Theorem~\ref{thm:anMlpIsGNM}} \label{app:ProofOfanMlpIsGNM}

Given a directed $K$-partite graph structure (with the convention of adding the bias node to every layer but the output nodes), its adjacency matrix is of the first-order off-(block)diagonal form:
\[ \mathbf{A} = 
    \begin{bmatrix}
    0 &  &  &  &  & \\
    \mathbf{M^{(1)}} & 0 &  &  &  &  \\
    0 & \mathbf{M^{(2)}} & 0 &  &  &  \\
     & 0 & \ddots & \ddots &  &  \\
     &  & \ddots & \ddots  & \ddots &\\
     &  &  & 0 & \mathbf{M^{(K)}} & 0
    \end{bmatrix}
\]
with 
\[
    \mathbf{M^{(k)}} = \begin{bmatrix} \mathbf{W}_k\ \textbf{b}_k \end{bmatrix}, \mathbf{W}_k \in \mathbb{R}^{n_k \times n_{k-1}}, \mathbf{b}_k \in \mathbb{R}^{n_k}\ 
 \forall k \in [K]
\]
and $\mathbf{A}_{ij}$ represents the edge weight of node $j$ to node $i$ ($j \rightarrow i$).

By asynchrony, only $\mathbf{M^{(t)}}$ is effective at time $t$, meaning that only nodes in the $t^{\text{th}}$ layer $v_{t,i}$ are updated from information contained in nodes $v_{t-1,j}$ from previous layer. 

\begin{equation*}
    \begin{aligned}
    h_{v_{t,i}}^{(t)} &= f \Big( \sum_{v_{t-1,j} \in \mathcal{N}^{\leftarrow}(v_{t,i})} \mathbf{M^{(t)}}_{ij} \, h_{v_{t-1,j}}^{(t-1)} \Big) \\
    &= f \Big( \mathbf{M^{(t)}} \cdot \mathbf{h_{v_{t-1,\cdot}}^{(t-1)}} \Big) \\
    &= f \Big( \mathbf{W}_t \mathbf{h_{z_{t-1}}^{(t-1)}} + \mathbf{b}_t \Big)\ 
    \end{aligned}
\end{equation*}
We thus recognize the layer update in an MLP of the corresponding $K$-partite structure as in (\ref{eq:MLPupdate}).

\subsection{Proof of Theorem~\ref{thm:anyMLPIsGNM}}
We prove the theorem by construction. To recover an MLP with $N$ non-bias neurons, we need a graph containing $N$ non-bias nodes plus a bias node, hence the total number of nodes equaling to $N+1$. For any valid number of layers $K$ such that $N = \sum_{k=1}^K n_k$, the parameter space of the GNM model applied on the $(N+1)$-dense graph is the set of tensors $\mathcal{A} \in \mathbb{R}^{K \times (N+1) \times (N+1)}$.  Let $\mathcal{A}_k$ denote the weight matrix of the $k^{\text{th}}$ update step, the conclusion comes from the trivial existence of the solution
\[
\mathcal{A}_k =
\begin{bmatrix}
    0 &  &  & & & & & 0 \\
    0 & \ddots &  & & & & & \vdots \\
    & \ddots&  &  & & & & \\
    & & 0  & \ddots&  & & & 0\\
    & & & \mathbf{W_k} & 0 & \cdots  & 0 & \mathbf{b}_k \\
    & & & & 0& \ddots &  &  0\\
   & & & & & \ddots & 0 & \vdots \\
   & & & & & & 0  &1  \\  
\end{bmatrix}
\]   
for every update at time $k \in [K]$, with $\mathbf{W}_k \in \mathbb{R}^{n_{k} \times n_{k-1}}$ and $\mathbf{b}_k \in \mathbb{R}^{n_k}$. $\mathbf{W}_k$ situates on the first order off-diagonal by-block. 
%Note that one can actually split the sub-matrices $\boldsymbol{M^{(k)}}$ (see the proof of the Theorem \ref{thm:anMlpIsGNM}) into the matrix $W_k$ and the column vector $b_k$. 
$\mathcal{A}_k$ is the effective adjacency matrix at time $k$, corresponding to the update of the nodes in the $k^{\text{th}}$ layer in an MLP.

\subsection{Proof of Corollary~\ref{cor:univ_approx}}
For all $N \in \mathbb{N}^*,\ \text{MLP}_{N,K} \subset \text{GNM}_{N+1,K} \subset C(\mathcal{K}\subset \mathbb{R}^m, \mathbb{R}^n)$: 
The first inclusion: By theorem \ref{thm:anyMLPIsGNM}, MLPs form a subset of GNM. The second inclusion: It is clear that a GNM is continuous since it is composed of continuous activation functions applied on linear transformations.

 Thus, the family of GNM contains the dense family (under $L^p, p \in [1,\infty]$ norms) MLP of continuous function \cite{hornik1989multilayer,leshno1993multilayer,arora2018understanding}, and is therefore itself a dense family in $C(\mathcal{K}\subset \mathbb{R}^m, \mathbb{R}^n)$.

\section{Dataset Statistics}\label{sec:datasets}
Table~\ref{tab:stats} gives the main statistics of the datasets used in this work.

\begin{table}[h]
    \centering
    \caption{Statistics of the datasets we used in our experiments.}
    \label{tab:stats}
    \footnotesize
    \renewcommand{\arraystretch}{1.2}
    \begin{tabular}{l|ccccc}
    \toprule
    Dataset & Task & \#samples & \#attributes & \#classes/targets \\
    \midrule
    Noisy Moons & Classification & 1,000 & 2 & 2 \\
    Abalone & Classification & 4,177 & 8 & 28 \\
    Adult & Classification & 48,842 & 14 & 2 \\
    Car & Classification & 1,728 & 6 & 4\\
    Connect4 & Classification & 67,557 & 42 & 3 \\
    Phishing & Classification & 11,055 & 30 & 2 \\
    Wireless & Classification & 2,000 & 7 & 4 \\
    Yeast & Classification & 1,484 & 8 & 10\\
    PageBlocks & Classification & 5,473 & 10 & 5 \\
    DryBean & Classification & 13,611 & 16 & 7 \\
    Letter Recognition & Classification & 20,000 & 16 & 26 \\
    Electrical Grid Stability & Classification & 10,000 & 12 & 2 \\
    Susy & Classification & 25,000 & 18 & 2 \\
    Isolet & Classification & 7,797 & 617 & 26\\
    Waveform & Classification & 5,000 & 40 & 3 \\
    \midrule
    Wine & Regression & 4,898 & 11 & 1 \\
    Auto MPG & Regression & 398 & 7 & 1\\
    Bike sharing & Regression & 17,389 & 13 & 1\\
    Diabetes & Regression & 442 & 10 & 1\\
    California & Regression & 20,640 & 8 & 1 \\
    Year & Regression & 515,345 & 90 & 1 \\
    \bottomrule
    \end{tabular}
\end{table}

\section{Computational Complexity}\label{sec:complexity}
\begin{figure}[h]
    \centering
    \includegraphics[width=.7\textwidth]{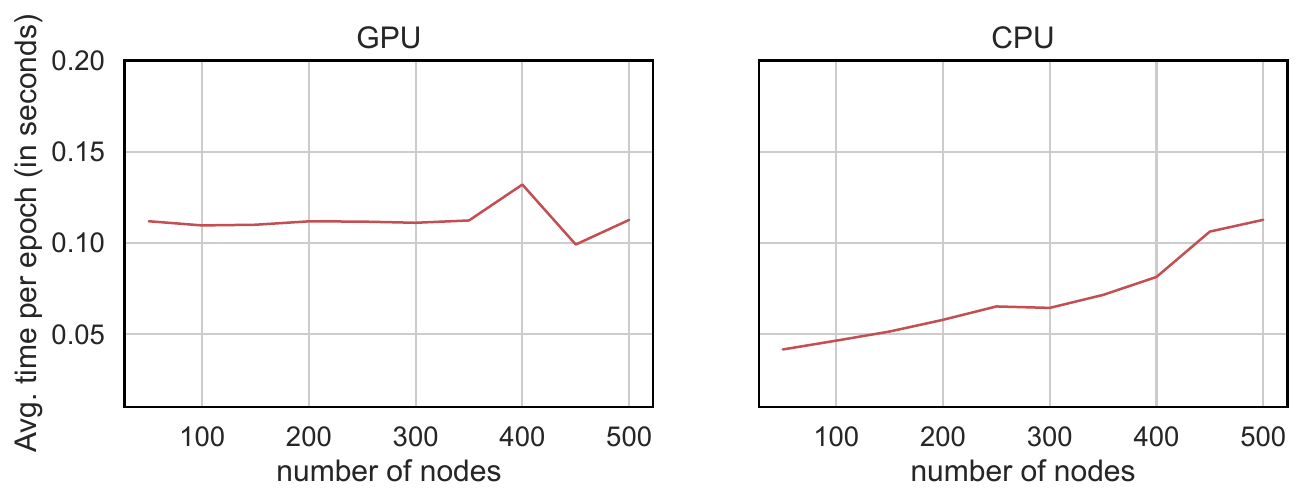}
    \caption{Average running time per epoch of the GNM model vs. number of nodes of $G_\text{GNM}$. Experiments performed on the Car dataset. Running time was measured on a NVIDIA RTX A6000 GPU (left) and on a Intel Xeon Silver 4215R CPU @ 3.20GHz (right).}
    \label{fig:runtime}
\end{figure}

To evaluate the runtime performance and scalability of the proposed GNM model, we measured how the model's average running time per epoch varies with respect to the number of nodes of graph $G_\text{GNM}$ in the Car dataset.
Figure~\ref{fig:runtime} illustrates the running time of GNM as a function of the number of nodes.
We have measured the running time both on a GPU (left) and on a CPU (right).
Note that in theory, running time grows quadratically with the number of nodes.
However, due to the parallel processing abilities of GPUs, we observe that the running time remains almost constant as the number of nodes increases.
In the case of GPU, there is indeed an increase in running time as the number of nodes increases.
However, it is not as large as expected by the theoretical analysis.

\begin{figure}[h]
    \centering
    \includegraphics[width=.7\textwidth]{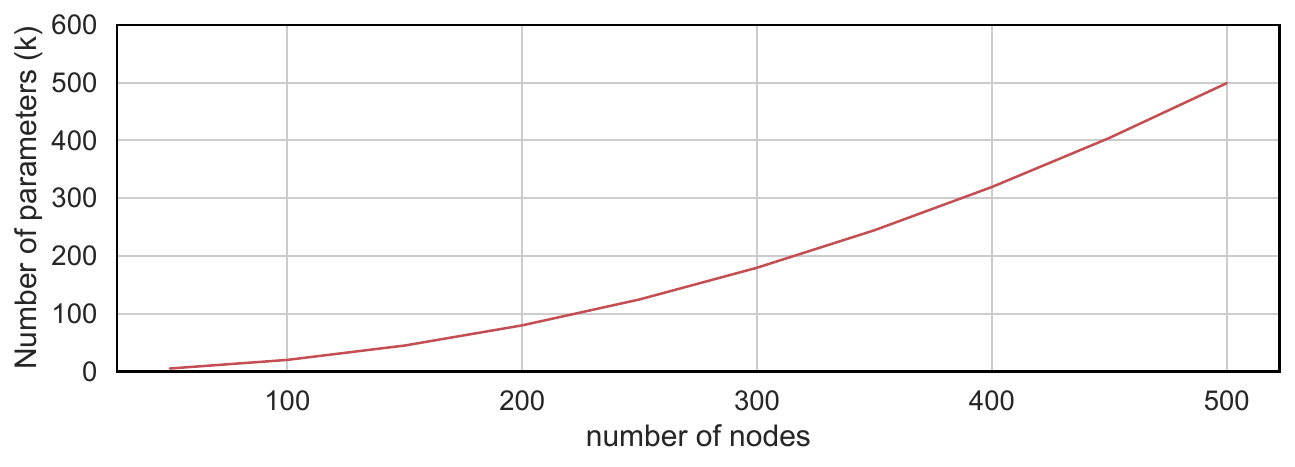}
    \caption{Number of parameters of the GNM model vs. number of nodes of $G_\text{GNM}$.}
    \label{fig:params}
\end{figure}

We have also measured how the number of parameters of the GNM model increases with respect to the number of nodes.
The results are shown in Figure~\ref{fig:params}.
We can see that there is a significant increase in the number of parameters as the number of nodes increases.
For example, when the number of nodes is equal to $50$, the model consists of $4.9$k parameters.
If we set the number of nodes to $500$, then the model consists of $499$k parameters.

\begin{figure}[h!]
    \centering
    \subfigure[]{\includegraphics[width=.4\textwidth]{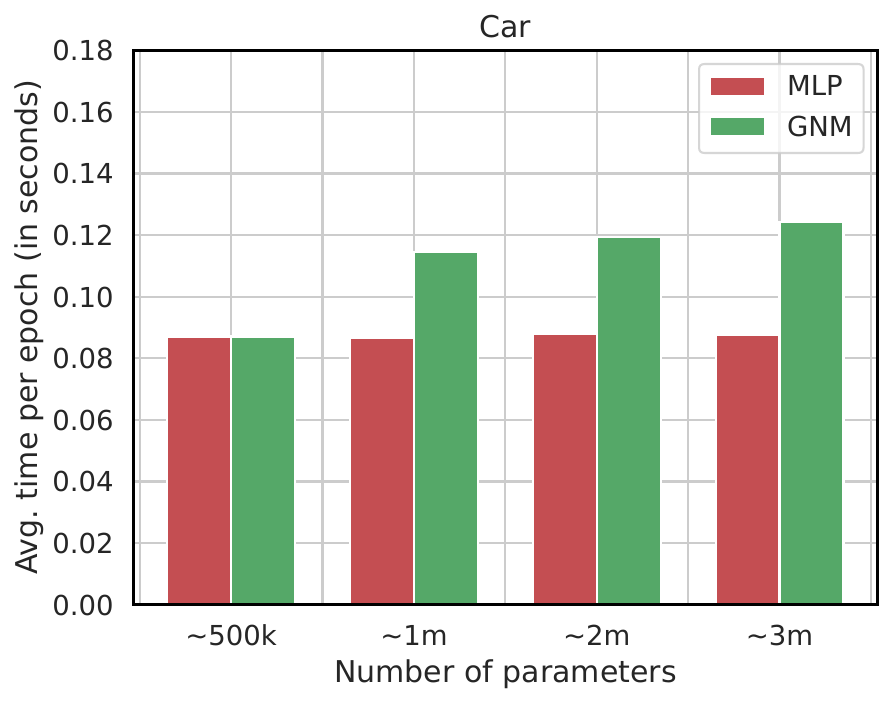}}\phantom{1cm}
    \subfigure[]{\includegraphics[width=.4\textwidth]{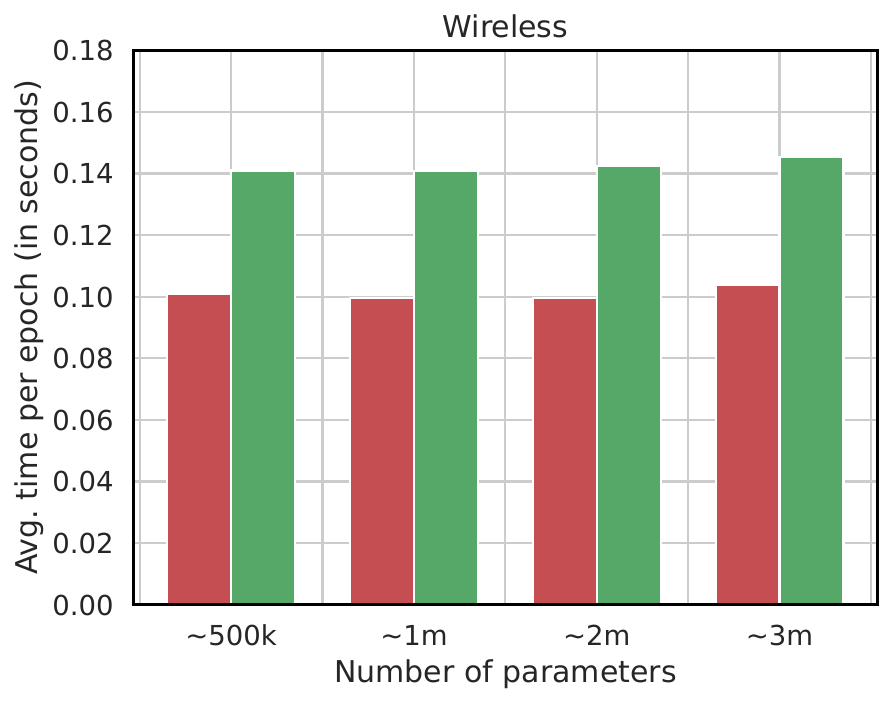}}
    \subfigure[]{\includegraphics[width=.4\textwidth]{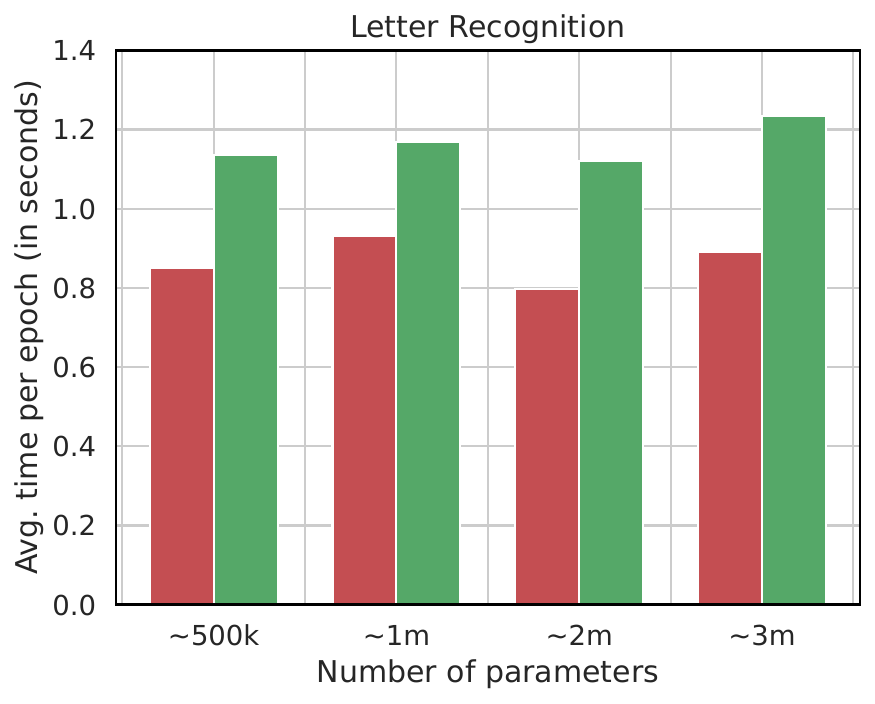}}\phantom{.1cm}
    \subfigure[]{\includegraphics[width=.4\textwidth]{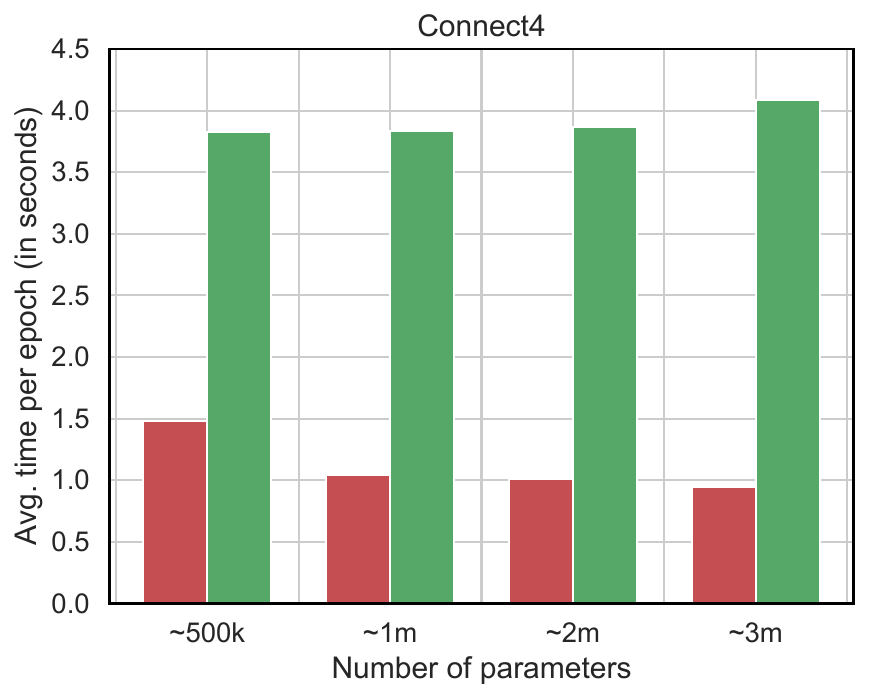}}
    \caption{Average running time per epoch of the GNM and MLP models on $4$ datasets with increasing number of parameters.}
    \label{fig:runtime_comparison}
\end{figure}

We also compare the proposed GNM model against MLP in terms of running time.
We set the number of layers of both models equal to $3$ and we also set the number of nodes of GNM and the hidden dimension size of MLP to values such that we obtain four different instances for each model where the number of parameters of the four instances are approximately equal to $500$k, $1$m, $2$m and $3$m parameters.
We train the different instances on a single split of the Car, Wireless, Letter Recognition and Connect4 datasets and we compute the average running time per epoch of each instance.
The results are illustrated in Figure~\ref{fig:runtime_comparison}.
Running times of all instances of both models were measured on a NVIDIA RTX A6000 GPU.
We observe that the MLP architecture is faster than the proposed GNM model on all four considered datasets.
However, the difference in average time per epoch is not very large.
Training the GNM model on the Car, Wireless and Letter Recognition datasets takes approximately $1.2 - 1.3$ times longer than training the MLP model.
The difference in running time is larger on the Connect4 dataset.
We should also mention that increasing the number of parameters usually has no impact on the running time of the two models.
For example, training a GNM model that consists of $3$m parameters takes approximately the same time as training a GNM model that consists of $500$k parameters.
The same was observed in Figure~\ref{fig:runtime}, and we attribute this to the parallel processing abilities of GPUs.

\section{Additional Training Curves}\label{sec:overfitting}
\begin{figure}[h]
    \centering
    \subfigure[Car]{\includegraphics[width=.49\textwidth]{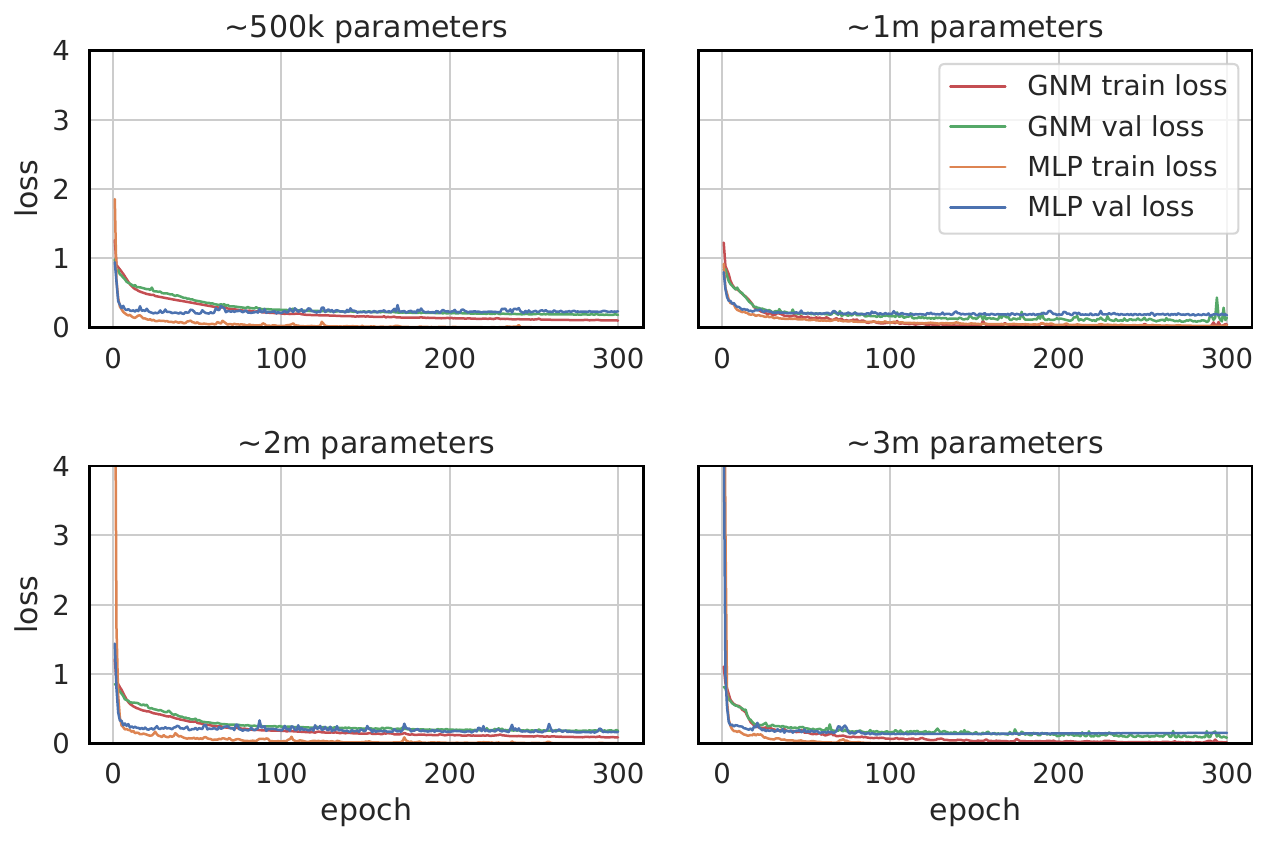}\label{fig:Ex_Im}}
    \subfigure[Phishing]{\includegraphics[width=.49\textwidth]{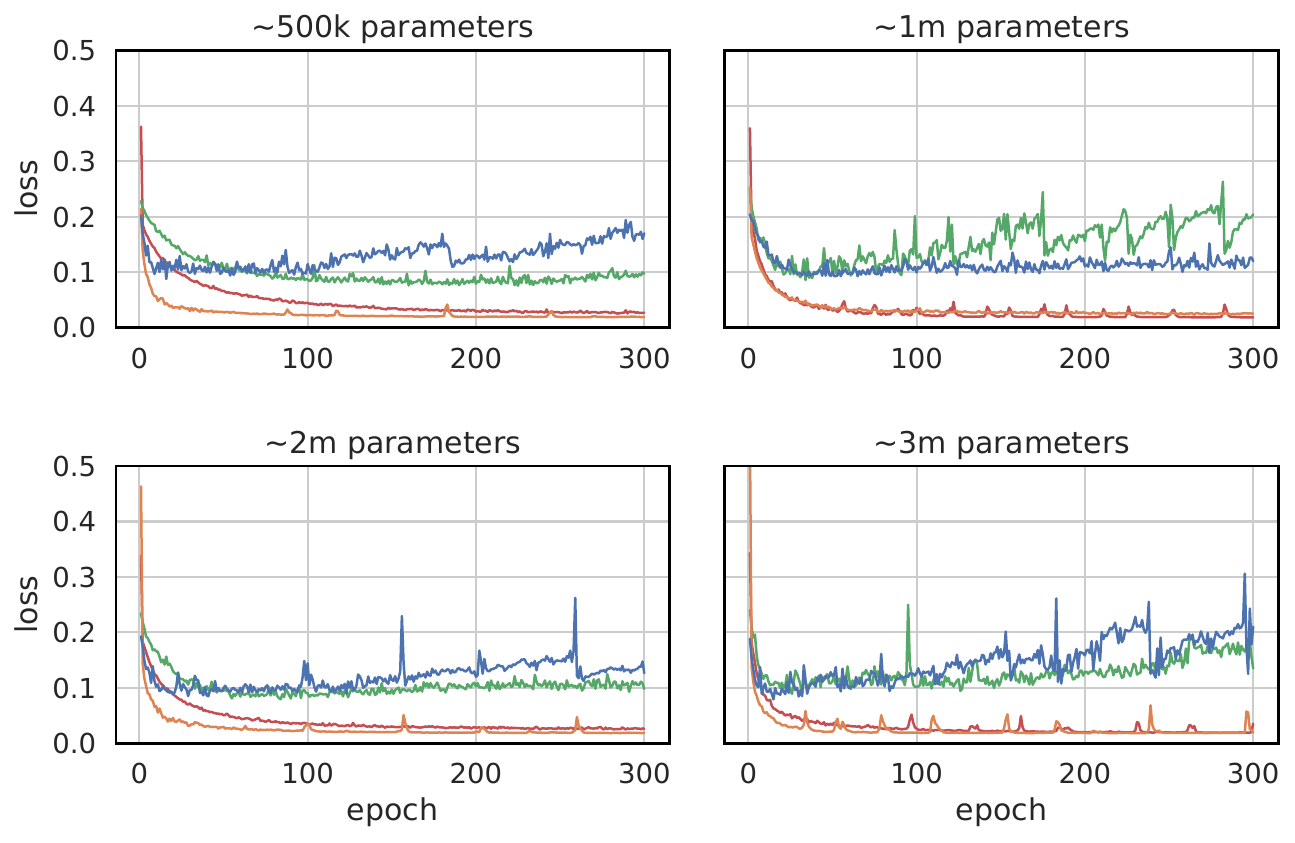}\label{fig:Ex_Im2}}
    \subfigure[Wireless]{\includegraphics[width=.49\textwidth]{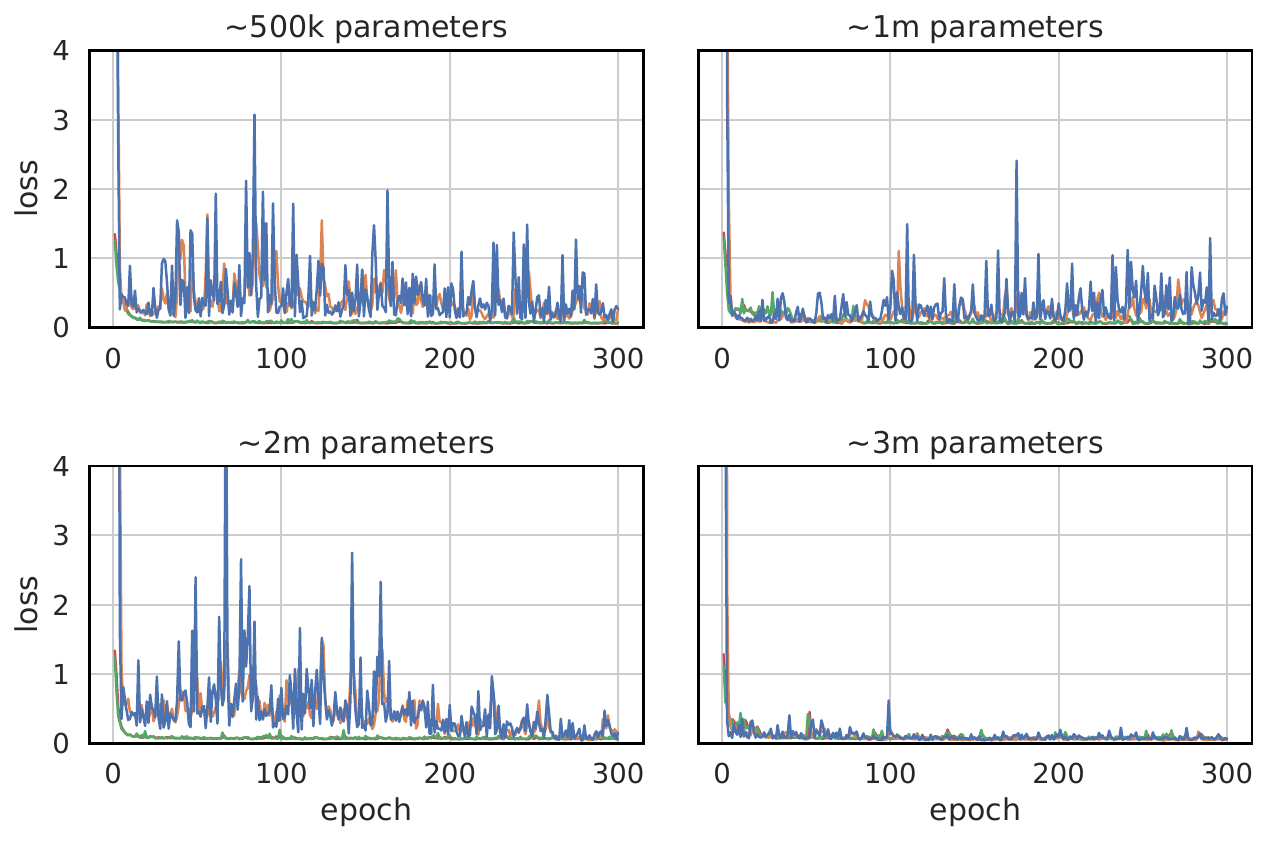}\label{fig:Ex_Im3}}
    \subfigure[Connect4]{\includegraphics[width=.49\textwidth]{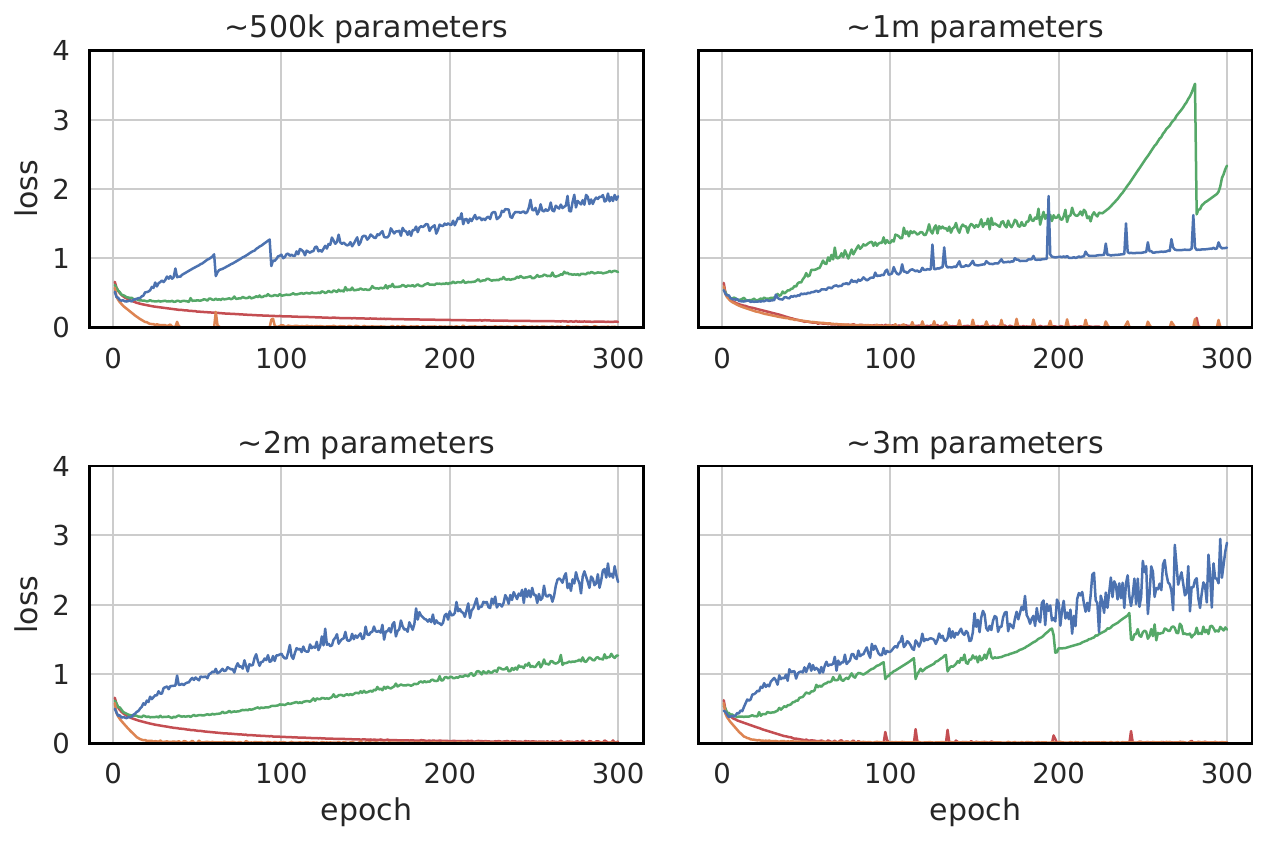}\label{fig:Ex_Im4}}
    \caption{Training curves of the GNM and MLP models on $4$ datasets. $4$ instances of each model are trained consisting of approximately $500$k, $1$m, $2$m and $3$m parameters.}
    \label{fig:training_curves}
\end{figure}

Figure~\ref{fig:training_curves} illustrates the training curves of the GNM and MLP models on $4$ datasets, namely Car, Phishing, Wireless and Connect4.
We can see that on Phishing and Connect4, both models suffer from overfitting, but GNM is generally less prone to overfitting than MLP.
On the Wireless dataset, we observe that the training procedure of MLP is somewhat unstable.
This is not the case for the GNM model.

\end{document}